\pdfoutput=1

\documentclass[11pt]{article}

\usepackage[final]{acl}
\usepackage{multirow}
\usepackage{times}
\usepackage{latexsym}
\usepackage{amsthm}
\usepackage{amsmath}
\usepackage{booktabs}
\usepackage{siunitx}

\newtheorem{lemma}{Lemma}[section]
\newtheorem{cor}{Corollary}[section]
\newtheorem{prop}{Proposition}[section]
\newtheorem{definition}{Definition}[section]

\usepackage[T1]{fontenc}

\usepackage[utf8]{inputenc}

\usepackage{microtype}

\usepackage{inconsolata}

\usepackage{graphicx}

\title{Local Normalization Distortion and the Thermodynamic Formalism of Decoding Strategies for Large Language Models}

\author{
 \textbf{Tom Kempton\textsuperscript{1, *}},
 \textbf{Stuart Burrell\textsuperscript{2, *}}
\\
\\
 \textsuperscript{1}Department of Mathematics, University of Manchester\\
 \textsuperscript{2}Innovation Lab, Featurespace\\
\\
 \small{
   \textbf{Correspondence:} \href{thomas.kempton@manchester.ac.uk}{thomas.kempton@manchester.ac.uk}
 }
}

\begin{document}

\maketitle
\begin{abstract}
Advances in hardware and language model architecture have spurred a revolution in natural language generation. However, autoregressive models compute probability distributions over next-token choices, and sampling from these distributions, known as decoding, has received significantly less attention than other design choices. Existing decoding strategies are largely based on heuristics, resulting in methods that are difficult to apply or improve in a principled manner. We develop the theory of decoding strategies for language models by expressing popular decoding algorithms as equilibrium states in the language of ergodic theory and stating the objective functions they optimize. Using this, we analyze the effect of the local normalization step required to make probabilities sum to one in top-k, nucleus, and temperature sampling. We argue that local normalization distortion is a fundamental defect of decoding strategies and quantify the size of this distortion and its effect on mathematical proxies for the quality and diversity of generated text. This yields conclusions for the design of decoding algorithms and the detection of machine-generated text.
\end{abstract}

\def\thefootnote{*}\footnotetext{Equal contribution.}\def\thefootnote{\arabic{footnote}}
\section{Introduction}
Autoregressive large-language models are poised to transform industries such as healthcare, finance and education \cite{zhao2024surveylargelanguagemodels}. Rapid advancements in this field have been fueled by scaling \cite{scaling}, transformer network architectures \cite{vaswani}, and alignment processes \cite{rlhf,dpo}. However, autoregressive language models produce conditional \emph{distributions} over predicted next tokens, and these must be iteratively sampled during inference. A suite of \emph{decoding} methods exist for this sampling, such as top-k, nucleus, or temperature sampling, but these are based on heuristics and our understanding of these methods is in its infancy, with most being developed through trial and error \cite{wiher2022decoding}.

This is surprising, since the choice of decoding strategy has a profound impact on the quality of the generated text and, in some cases, may be more important than the model architecture \cite{wiher2022decoding}. For example, greedy sampling, or the related concept of beam search, tends to produce accurate but repetitive or dull text \cite{holtzmancurious}. In contrast, pure sampling produces much more varied and interesting text but can produce text which is `incoherent and almost unrelated to the context' \cite{holtzmancurious}. 

Our primary goal is to fill this gap and initiate the theoretical study of decoding strategies. As an application of this theory, we investigate a well-known defect of popular decoding strategies, which results from truncating the distribution of possible next tokens at each step during inference. Truncation necessitates repeated renormalization to obtain valid probability distributions, and results in a phenomenon we term \emph{local normalization distortion}. The fact that local normalization distorts the resulting probability distribution is well-known, but not well-understood. 

Our contributions are as follows.
\begin{enumerate}
\item Develop a theoretical framework for analyzing decoding strategies. This involves describing the probability distributions $q$ produced by decoding algorithms as equilibrium states, drawing heavily on the language of ergodic theory and thermodynamic formalism \cite{bowen2008equilibrium}. We precisely state the objective function maximized by each decoding strategy; see Section \ref{sec:Equilibrium States}.
\item Quantify the effect of local normalization distortion by showing how the probability of randomly chosen strings change when we replace a locally normalized decoding strategy with a globally normalized equivalent. The effect is large for top-k and temperature sampling, but much smaller for nucleus sampling; see Section \ref{sec:experiment6}.
\item Evaluate language models and decoding strategies in terms of a quality-diversity trade-off, as in \newcite{caccia2019}. We show, both theoretically and empirically, that local normalization distortion negatively affects the performance of the decoding strategy.\footnote{Imperfect mathematical proxies for quality and diversity of text are used in this analysis, see Section \ref{sec:evaluating}. Code to reproduce our experiments is available at \href{https://github.com/TMKempton/lnd}{https://github.com/TMKempton/lnd}.}
\end{enumerate}
These results show that in the ongoing search for better decoding strategies for large language models, careful attention should be paid to local normalization distortion, as it may have a large effect on proxies for the quality of generated text, and the size of this effect varies considerably with choice of decoding strategy. Additionally, in a follow-up article \cite{kempton2025temptest} we show how to use local normalization distortion to detect machine-generated text.





\section{A Motivating Example}
Many autoregressive models for natural language generation work broadly as follows. Given a vocabulary $\mathcal V$, one builds a large neural network to estimate the likelihood $p(y_t|{\bf y_{<t}})$ that the next token in a sequence is equal to $y_t\in\mathcal V$, given the previous tokens ${\bf y_{<t}}=y_{0}\cdots y_{t-1}\in\mathcal V^{t}$. Then, one decides on a decoding strategy (sampling algorithm), which is a way of using the collection of likelihoods
\[
\{p(y_t|{\bf y_{<t}}):y_t\in\mathcal V\}
\]
associated with context ${\bf y_{<t}}$ to choose the next token $y_t$. For example, one could always choose the token $y_t$ which has highest likelihood $p(y_t|{\bf y_{<t}})$ (greedy sampling), or one could allow each token $y_t$ to be chosen with probability equal to the likelihood $p(y_t|{\bf y_{<t}})$ (pure sampling). Let $q(\cdot|{\bf y_{<t}})$ denote the distribution of choices of $y_t$ given the context and chosen decoding strategy.

Having chosen the token $y_t$, one repeats the process to choose $y_{t+1}$ given the new context ${\bf y_{<t+1}}$. One computes the probability of a string $y_0\cdots y_T\in\mathcal V^*$ by
\[
q(y_0\cdots y_T)=\prod_{t=1}^Tq(y_t|{\bf y_{<t}}).
\]


In many settings, rather than conditioning on the whole history ${\bf y_{<t}}$ one allows only a finite context length $L$ and conditions on $y_{t-L}\cdots y_{t-1}$, making the process of generating texts an $L$-step Markov process.

\begin{figure}
    \centering
\includegraphics[width=\linewidth]{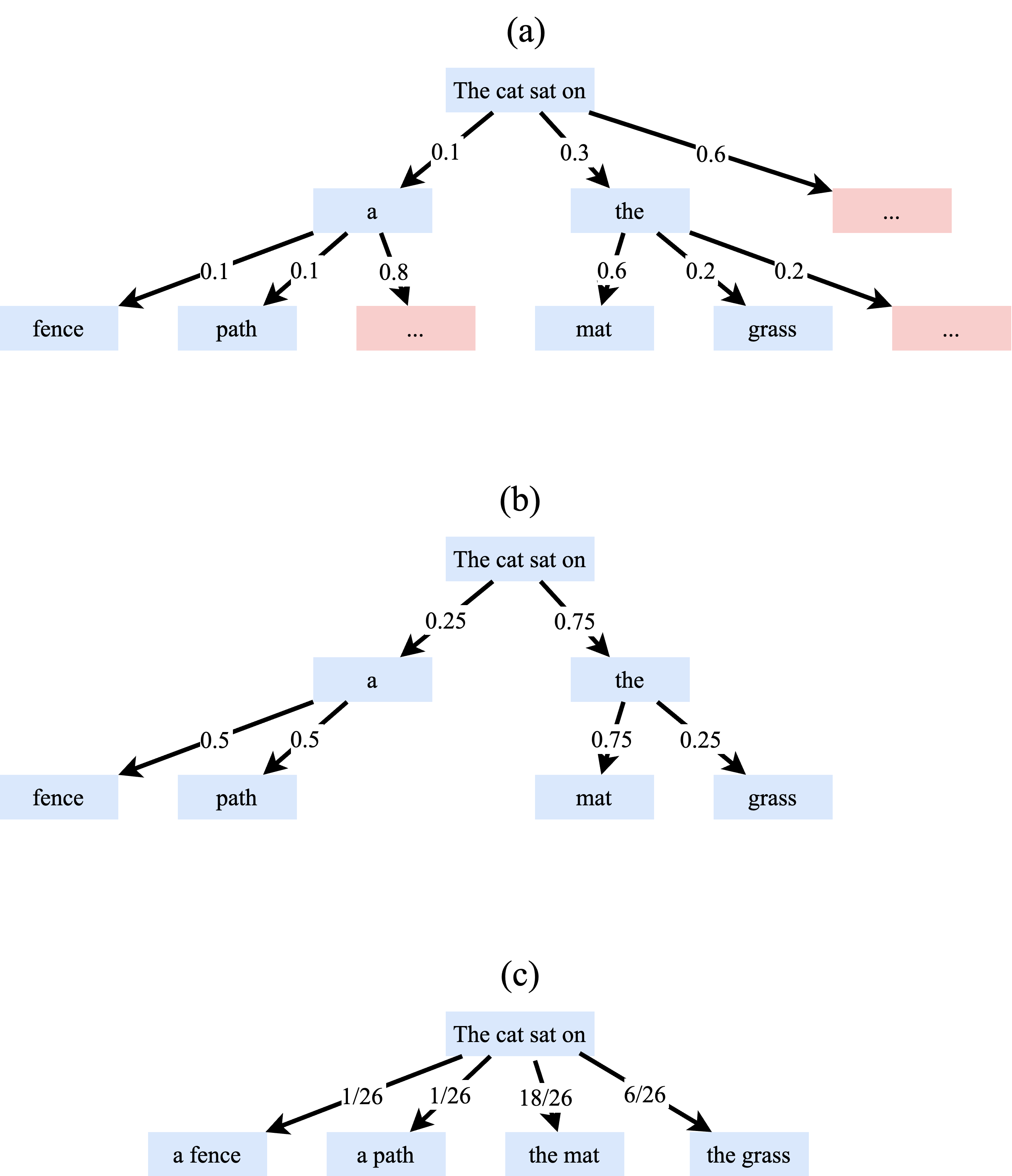}
\caption{Distortion due to local normalization significantly impacts the implied probability distribution during decoding. (a) shows pure sampling, (b) shows locally normalized sampling, and (c) shows globally normalized sampling.}
    \label{ToyExample}
\end{figure}

A simple example shows how the decoding strategy can dramatically influence $q$. Suppose we have a language model that produces model likelihoods $p$. We feed in the context `The cat sat on' and obtain the likelihoods of the various choices of the next two tokens. Let these likelihoods be as described in Figure \ref{ToyExample}. Assume further that we decide that various choices of tokens are unreliable and so we wish to restrict our choices to the two most likely tokens depicted in blue boxes. We now have probabilities which do not sum to one, and so have to decide how to normalize them. 

\paragraph{Option 1: Global normalization.} Compute the probabilities of complete strings and divide by the sum of the probabilities of complete strings. In our example, the sum of the probabilities of complete strings is $0.26=0.01+0.01+0.18+0.06$. We end up with `The cat sat on a fence' being selected with probability $0.01/0.26\approx 0.038$.

\paragraph{Option 2: Local normalization.} Renormalize conditional probabilities locally so that the probabilities of outward edges from each node sum to one. For example, we choose the first word `a' with probability $0.1/0.4$ and `the' with probability $0.3/0.4$. With local normalization, we end up with `The cat sat on a fence' being selected with probability $0.25\times 0.5=0.125$. 
\newline\\
Observe that these two different methods of normalizing probabilities yield very different probability distributions.

\section{Decoding Strategies}
The decoding strategies we study in this article fall into two classes, truncation sampling algorithms (such as nucleus sampling and top-k) and temperature sampling. Truncation sampling algorithms work by defining an allowed set $\mathcal A_{\bf y_{<t}}$ of tokens that can follow some context, restricting the model conditional probability distribution $p(\cdot|\bf y_{<t})$ to this set and renormalizing it to have mass one.

\begin{definition}[Truncation Sampling Algorithms]
Given a language model $p$, a context ${\bf y_{<t}}$ and an allowed set $\mathcal A_{\bf y_{<t}}$, define \[Z({\bf y_{<t}})=\sum_{w_t\in \mathcal A_{\bf y_{<t}}}p(w_t|{\bf y_{<t}}).\]
Choose element $y_t$ of $\mathcal A_{\bf y_{<t}}$  with probability \[q(y_t|{\bf y_{<t}}):=\dfrac{p(y_t|{\bf y_{<t}})}{Z({\bf y_{<t}})}\]
and set
\[q(y_t|{\bf y_{<t}})=0
\]for $y_t\not\in \mathcal A_{\bf y_{<t}}$.
\end{definition}
Top-k sampling \cite{fan2018hierarchical} is an example of a truncation sampling strategy. Given some value of $k\geq 1$, it is defined by setting the allowed set $\mathcal A_{\bf y_{<t}}$ to be the set of those $k$ tokens with highest model probabilities $p(y_t|{\bf y_{<t}})$. Top-k sampling restricts to the $k$ most likely tokens at each stage of language generation, and renormalizes mass at each stage of generation by dividing model probabilities by the sum of the probabilities of the top-k tokens.

Nucleus (top-p) sampling \cite{holtzmancurious} is defined by choosing a value $\pi\in[0,1]$, ordering tokens $w_1,w_2\cdots \in\mathcal V$ in order of decreasing model probability $p(w_i|{\bf y_{<t}})$ and setting $\mathcal A_{\bf y_{<t}}=\{w_1,\cdots, w_r\}$ where the threshold $r$ is the smallest natural number satisfying $\sum_{i=1}^r p(w_i|{\bf y_{<t}})\geq \pi$. Rather than choosing from exactly $k$ tokens at each stage, nucleus sampling samples from (roughly) the top proportion $\pi$ of the probability distribution. It is worth stressing that  $\sum_{i=1}^r p(w_i|{\bf y_{<t}})$ is often much larger than $\pi$, and so the normalizing constant $Z({\bf y_{<t}})=\sum_{i=1}^r p(w_i|{\bf y_{<t}})$ can vary significantly at different contexts.

Several newer truncation-based decoding strategies have been introduced with different clever ways of choosing the allowed sets $\mathcal A_{\bf y_{<t}}$, see, for example, locally typical sampling \cite{meister2023locally}, $\eta$-sampling \cite{hewitt2022truncation}, basis-aware truncation \cite{finlayson2023closing} and microstat \cite{basu2020mirostat}. One observation about this body of research is that careful motivation is always given for the choice of allowed set, but little if any consideration is given to the way in which the resulting decoding strategy apportions probability mass within this allowed set.

Finally, we mention temperature sampling, which is the only widely used stochastic sampling algorithm not to fall into the framework of truncation sampling.
\begin{definition}[Temperature Sampling \cite{guo2017calibration}]
Given some context ${\bf y_{<t}}$ and a parameter $\tau>0$ (usually $\tau\in(0,1))$, define
\[Z_{\tau}({\bf y_{<t}})=\sum_{w_t\in \mathcal V}(p(w_t|{\bf y_{<t}})^\frac{1}{\tau}).
\]
The distribution $q_{\tau}$ given by temperature sampling is defined by
\[
q_{\tau}(y_t|{\bf y_{<t}})=\dfrac{p(y_t|{\bf y_{<t}})^\frac{1}{\tau}}{Z_{\tau}({\bf y_{<t}})}.
\]

\end{definition}

\subsection{Global Normalization}\label{sec:globalnorm}
As in our toy example, we could replace the local normalization in top-k, nucleus and temperature sampling with a global normalization, in which rather than normalizing conditional probabilities by dividing by $Z({\bf y_{<m+i}})$ at each step, one normalizes the joint distribution over complete sequences $w_1\cdots w_T$. For example, if we let ${\bf \mathcal A_T}$ denote the set of sequences of length $T$ for which each token is in the top-k set, globally normalized top-k sampling selects string $y_1\cdots y_T\in $ with probability
\begin{equation}\label{eq:global_top_k}
q_k'(y_1\cdots y_T)=\dfrac{p(y_1\cdots y_T)}{\sum_{w_1\cdots w_T\in {\bf \mathcal A_T}} p(w_1\cdots w_T)}.
\end{equation}

That is, globally normalized top-k sampling samples according to the measure $p$ conditioned on the subset ${\bf\mathcal A_T}$. Similar statements hold for globally normalized variants of any restriction sampling algorithm. Globally normalized temperature sampling selects tokens with probability proportional to $p(\cdot|{\bf y_{<t}})^{\frac{1}{\tau}}$, as is the case when temperature is used in statistical physics, ergodic theory and fractal geometry (see Appendix \ref{sec:gibbsmeasures}).

We let $q'_k$, $q'_{\pi}$ and $q'_{\tau}$ denote the globally normalized alternatives to top-k, nucleus and temperature sampling respectively. Globally normalized decoding strategies are computationally infeasible, even for fairly small values of $T$. We introduce them here as a theoretical tool to better understand how problematic local normalization distortion is. In Appendix \ref{app:rejection} we explain how to sample from $q'_k, q'_{\pi}$ and $q'_{\tau}$ using rejection sampling.

\subsection{Local Normalization Distortion}
\begin{definition}\label{def:LND}
Let $q$ be the distribution produced by a locally normalized decoding strategy, and let $q'$ be its globally normalized counterpart. Given a context ${\bf y_{<t}}$, the local normalization distortion associated to completion $y_t\cdots y_T$ is defined as
\[
\dfrac{q(y_t\cdots y_T|{\bf y_{<t}})}{q'(y_t\cdots y_T|{\bf y_{<t}})}.
\]
\end{definition}

In the case of top-k sampling, given context ${\bf y_{<t}}$, there is a constant $C$ such that each completion $y_t\cdots y_T$ has local normalization distortion
\[
\dfrac{1}{C}\cdot\dfrac{1}{\prod_{i=0}^{T-t} Z_k({\bf y_{<t+i}})},
\]
where $Z_k$ is the mass of the top-k tokens at context ${\bf y_{<t+i}}$. The constant $C$ is the normalizing constant associated to global normalization, which is hard to compute but can be bypassed in empirical investigations, see Section \ref{sec:experiments}.

There is a body of work studying global normalization in the context of constrained decoding; see, for example, \newcite{lipkin2025fastcontrolledgenerationlanguage,loula2025syntactic} and references therein. In these works, local normalization distortion is seen as a problem and various algorithms for sampling approximately from the globally normalized distribution are proposed.

By contrast, \newcite{gareev-etal-2024-local} argue that global decoding underperforms local decoding for top-k and top-$\pi$ sampling. The primary driver of the effects they observe seems to be the fact that local and global sampling produce texts of markedly different lengths (differing by a factor of four for some parameter settings), whereas we study generations of fixed length. Additionally, their approach differs from ours in that they compare quality of local and global top-k sampling at the equal values of k, whereas we compare quality of local and global top-k sampling at equal values of diversity, following the approach of \newcite{caccia2019}. We justify our experimental approach in Section \ref{sec:experiments} and give further details on the difference between our work and \newcite{gareev-etal-2024-local} in Appendix \ref{sec:Gareev}. 

We conclude this section with a proposition that further motivates the study of local normalization distortion.

\begin{prop}\label{prop:zerotemp}
Let $p$ be a language model, $q_{\tau}$ denote the distribution arising from temperature sampling, and $q'_{\tau}$ the distribution arising from a globally normalized version of temperature sampling. Then, as the temperature parameter $\tau$ tends to zero, $q_{\tau}$ converges to the distribution putting all of its mass on the output of greedy sampling, whereas $q'_{\tau}$ converges to the distribution putting all of its mass on the sequence with globally maximal $\log$ probability. 
\end{prop}
\begin{proof}
    See Appendix \ref{proof:zerotemp}.
\end{proof}
Given the substantial interest in implementing expensive search algorithms such as beam search to find sequences with approximately the globally maximal log probability, we present Proposition \ref{prop:zerotemp} as initial evidence that local normalization distortion can have a substantial negative effect and is worthy of further investigation.

\section{Evaluating Decoding Strategies through a  Quality-Diversity Trade Off}\label{sec:evaluating}
When generating text from a language model, one may have different preferences for the `quality' and `diversity' of the text according to the task being performed \cite{caccia2019, wiher2022decoding}.  By diversity, we mean the capacity for a model to produce different samples, while by quality we mean the average human judged quality of an individual sample. 

This framing of choice of decoding strategy choice as a trade-off of quality and diversity is studied in \newcite{caccia2019, ippolito2019comparison,nadeem2020systematic, zhang2021trading}. In particular, in \newcite{caccia2019}, the authors propose using a `temperature sweep' to find a parameter $\tau$ for which temperature sampling best matches this preference. Similarly, one can adjust the parameters in top-k or nucleus sampling according to one's preferences for diversity versus quality, since restricting token choices to the top of the distribution prioritizes quality over diversity.

There are no universally accepted definitions of the diversity and quality of text. One way of evaluating the diversity of stochastically generated text is to look at the entropy $H(q)$ of the distribution $q$ of the text, given by
\[
H(q)=\sum_{{\bf y}\in \mathcal V^{*}}q({\bf y})\log q({\bf y}).
\]
The sum here is taken over complete strings.

The gold standard for evaluating the quality of the generated text is to get human judgment scores, although this is expensive and fraught with difficulty \cite{clark2021all}. Often the model log-likelihood $\log(p)$ is used as a proxy for quality, so the quality of a distribution $q$ over possible texts would be given by

\[
Q(q)=\sum_{{\bf y}\in \mathcal V^{*}}q({\bf y})\log p({\bf y}).
\]
This notion of quality is not without its issues \cite{ meister2022probability}, although \newcite{zhang2021trading} has a rather compelling graph suggesting that human judgements of quality of text are well correlated with $Q(q)$ except when $Q(q)$ is very high. 

While entropy and average log-likelihood of a distribution $q$ are imperfect, albeit frequently used, proxies for diversity and quality, they are precisely the right objects to describe mathematically the objective functions maximized by the distributions resulting from top-k sampling, nucleus sampling and temperature sampling.

\section{Decoding Strategies as Equilibrium States}\label{sec:Equilibrium States}
In the last section we reviewed the literature on what a decoding strategy ought to maximize. In this section we prove results about what popular decoding strategies actually optimize. In particular, we state results of the form `given some context ${\bf y_{<m}}$, the probability distribution $q$ on the set $\mathcal A^*_{{\bf y_{<m}},k}$ obtained by sampling according to a certain decoding strategy is the unique probability distribution on $\mathcal A^*_{{\bf y_{<m}},k}$ that maximizes the following objective function...'.

In the language of ergodic theory, what we are doing is describing the outcome of a decoding strategy as an {\it equilibrium state} associated to a certain potential. While the mathematics of this section is not hard, it is useful as it allows us to ask whether the function that our decoding strategy maximizes is well aligned with the theoretical goals of a decoding strategy. In a follow-up paper \citep{kempton2025temptest} we apply the ideas of this section to build a novel algorithm for detecting text generated by a language model.

We use the following standard result. As usual, let $0\log 0:=0$. 
\begin{lemma}[\cite{bowen2008equilibrium}, Lemma 1.1]\label{lemma:bowen}
Let $X=\{1,\cdots,k\}$ be a finite set and let $R=(r_1,\cdots,r_k)$ be a probability measure on $X$ assigning mass $r_i$ to symbol $i$. Then $R$ is the unique probability measure maximizing the quantity
\[
\underbrace{-\sum_{i\in X}\mu_i\log \mu_i}_{\mbox{ Entropy }H(\mu)}+\underbrace{\sum_{i\in X}\mu_i\log r_i}_{\mbox{Average log probability}}
\]
among probability measures $\mu=(\mu_1,\cdots \mu_k)$ on $X$.
\end{lemma}
The results of this section follow as direct corollaries to Lemma \ref{lemma:bowen} by analyzing the measures $(r_1,...,r_k)$ given by various decoding strategies. Details are given in the appendix \ref{sec:proofs}. Our first result concerns top-k decoding.
\begin{cor}\label{topkeqstate}
Given some context ${\bf y_{<m}}$ and a choice of $k$, the distribution $q_k$ on $\mathcal A^*_{{\bf y_{<m}},k}$ produced by top-k sampling is the unique distribution maximizing the quantity
\begin{eqnarray*}\label{topkmax}
& &\underbrace{H(\mu)}_{\mbox{ Proxy for diversity}}\\&+&\underbrace{\sum_{{\bf w}\in \mathcal A^*_{{\bf y_{<m}},k}} \mu({\bf w|y_{<m}})\log p({\bf w|y_{<m}})}_{\mbox{Proxy for quality}}\\&+&\underbrace{\sum_{{\bf w}\in \mathcal A^*_{{\bf y_{<m}},k}} \mu({\bf w|y_{<m}})\log \epsilon_k({\bf w})}_{\mbox{Distortion term}}
\end{eqnarray*}
among distributions $\mu$ on $\mathcal A^*_{{\bf y_{<m}},k}$. Here
\[
\epsilon_k({{\bf w}}):=\frac{1}{\Pi_{i=0}^{T-m} Z({ y_0\cdots y_mw_{m+1}\cdots w_{m+i-1}})},
\]
which is the product along the sequence of the inverse of the mass of the top k tokens.
\end{cor}
If our equation above had only the first two terms, it would show that $q_k$ maximizes the sum of mathematical proxies for diversity and quality among distributions supported on $\mathcal A^*_{{\bf y_{<m}},k}$. The third term however, which is an artifact of local normalization, distorts this goal. If the third term was constant across sequences ${\bf y}$ then it would have no effect, but 
issues arise when it has a large variance; see Section \ref{sec:experiment6}. Next, we consider nucleus sampling.

\begin{cor}\label{toppeqstate}
Given some context ${\bf y_{<m}}$ and a choice of $\pi$, the distribution $q_{\pi}$ on $\mathcal A^*_{{\bf y_{<m}},\pi}$ generated by nucleus sampling is the unique distribution maximizing the quantity
\begin{eqnarray*}\label{toppmax}
H(\mu)&+&\sum_{{\bf w}\in \mathcal A^*_{{\bf y_{<m}},\pi}} \mu({\bf w|y_{<m}})\log p({\bf w|y_{<m}})\\
&+&\sum_{{\bf w}\in \mathcal A^*_{{\bf y_{<m}},\pi}} \mu({\bf w|y_{<m}})\log \epsilon_{\pi}({\bf w})
\end{eqnarray*}
among distributions $\mu$ on $\mathcal A^*_{{\bf y_{<m}},\pi}$. Here
\[
\epsilon_{\pi}({{\bf w}}):=\frac{1}{\Pi_{i=0}^{T-m} Z({ y_0\cdots y_mw_{m+1}\cdots w_{m+i-1}})},
\]
which is the inverse of the product along the sequence $w_{m+1}\cdots w_T$ of the total mass of those tokens allowed by nucleus sampling. \end{cor}
Thus nucleus sampling produces a distribution $q_{\pi}$ maximizing a goal related to quality, diversity and an error term related to both by the length of the sequence $y$ and the extent to which the mass of the tokens selected by nucleus sampling overshoots the target $\pi$. 

Finally, we consider temperature sampling.

\begin{cor}\label{tempeqstate}
Given some choice of temperature $\tau$ and context ${\bf y_{<m}}$, the distribution $q_{\tau}$ is the distribution maximizing the quantity
\begin{eqnarray*}\label{maineq}
H(\mu)+\frac{1}{\tau}\sum_{{\bf w}\in \mathcal V^*} \mu(y_1\cdots y_m{\bf w})\log(p(y_1\cdots y_m{\bf w}))\\
+\sum_{{\bf w}\in \mathcal V^*} \mu(y_1\cdots y_m{\bf w})\epsilon_{\tau}({\bf w})
\end{eqnarray*}
among distributions $\mu$ on $\mathcal V^*$, where
\[
\epsilon_{\tau}({\bf w})=\frac{1}{\Pi_{i=0}^{T-m} Z_{\tau}({ y_0\cdots y_mw_{m+1}\cdots w_{m+i-1}})}. 
\]
\end{cor}

When nucleus or top-k sampling set $\pi<1$ or $k<|\mathcal V|$, they produce distributions with lower entropy than $p$, since token choice has been reduced. This process also redistributes mass from low probability tokens to higher probability tokens, reducing the average log-likelihood. Thus the parameters associated with nucleus or top-k sampling allow one to prioritize quality or diversity. This is somewhat hidden in the statements of Corollaries \ref{topkeqstate} and \ref{toppeqstate}, it appears only in that it is the space of distributions on $\mathcal A^*$ upon which the proxy for diversity, proxy for quality, and distortion term are maximized.  

Temperature sampling is not a truncation algorithm, and the way it modulates between prioritizing quality and prioritizing diversity is in the factor ${1}/{\tau}$ preceding the mathematical proxy for quality in the main equation of Corollary \ref{tempeqstate}. As before, the third term here is not related to the goal of maximizing quality or diversity. It is broadly similar to the distortion introduced by top-k, except rather than dividing by the product along a sequence of the mass contained in the top-k tokens, it divides by the product along a sequence $y_m\cdots y_T$ of the $L_{1/\tau}$ norm of the distribution $p(\cdot|{\bf y_{<m+i}})$.

Finally, we see that global normalization removes the problematic third term in each of the quantities maximized by our decoding strategies. 
\begin{cor}\label{globaleqstate}
In each of corollaries \ref{topkeqstate}-\ref{tempeqstate}, if the locally normalized probability distribution $(q_k,$ $ q_{\pi}$, or $ q_{\tau})$ is replaced by its globally normalized equivalent, the statement of the theorem remains the same except without the third term (i.e. the distortion term) in the expression for the maximized quantity.
\end{cor}
This is important because it shows that, in terms of the quality-diversity trade off goals of the previous section, locally normalized decoding strategies perform strictly worse than their globally normalized counterparts. In Section \ref{sec:experiment7} we quantify this effect.

\section{Experiments}\label{sec:experiments}
In this section we quantify the size of effects predicted in our theoretical sections. In particular, we measure the size of local normalization distortion and its effect on the quality-diversity trade off, measured through log-likelihood and entropy. 

While one may, in some circumstances, care more about semantic diversity than lexical diversity, or about human judgements of quality than language model log-likelihood, effects on these metrics would be downstream of the true mathematical effect of local normalization distortion, and likely highly dependent on both the language model and the task being performed. Thus, for reasons both of stability and of quantifying our theoretical results directly, we run our experiments on metrics present in the objective functions maximized by decoding strategies.

The experiments of this section are run using Llama 2-7B \cite{touvron2023llama} on a single A100 GPU. Detailed setup for each experiment is contained in the corresponding sections and appendices. In Appendix \ref{supp:app}, we repeat our experiments using Pythia 1B and 2.8B \cite{BidermanPythia}, and Llama 3.2 1B and 3.2 3B \cite{grattafiori2024llama}. As predicted above, our experimental results remain qualitatively unchanged.

The primary challenge to running these experiments is the computational cost of global normalization. To do this efficiently, we introduce a process based on rejection sampling in Appendix \ref{app:rejection}.

\subsection{How Large is the Distortion due to Local Normalization under Different Decoding Strategies?}\label{sec:experiment6}

In assessing how much local normalization distortion affects the mass of a completion $y_m\cdots y_T$, we need to compare how much  $q(y_m\cdots y_T|{\bf y_{<m}})$ is boosted by local normalization against how much it would have been boosted by global normalization, as in Definition \ref{def:LND}. Considering, for example, top-k sampling, if $q_k$ denotes the distribution produced by top-k sampling and $q_k'$ denotes its globally normalized equivalent, we would like to compute \[\dfrac{q_k(y_m\cdots y_T|{\bf y_{<m}})}{q_k'(y_m\cdots y_T|{\bf y_{<m}})}.\]
This is difficult to compute because globally normalized top-k assigns mass
\[
q_k'(y_m\cdots y_T|{\bf y_{<m}})=\dfrac{p(y_m\cdots y_T|{\bf y_{<m}})}{C}
\]
for some constant $C$, which is very expensive to compute, particularly on long generations. Instead, we generate pairs of completions $y_m\cdots y_T$ and $z_m\cdots z_T$ by top-k sampling and then compute the ratio of the two local normalization distortions by computing
\begin{eqnarray}\label{eq:LNDRatio}
   \dfrac{q_k(y_m\cdots y_T|{\bf y_{<m}})}{q_k'(y_m\cdots y_T|{\bf y_{<m}})}\large / \dfrac{q_k(z_m\cdots z_T|{\bf y_{<m}})}{q_k'(z_m\cdots z_T|{\bf y_{<m}})}
    \end{eqnarray}
    which is equal to
    \begin{eqnarray}\label{ratioLND}
  \dfrac{q_k(y_m\cdots y_T|{\bf y_{<m}})}{p(y_m\cdots y_T|{\bf y_{<m}})}. \dfrac{p(z_m\cdots z_T|{\bf y_{<m}})}{q_k(z_m\cdots z_T|{\bf y_{<m}})}.
\end{eqnarray}
Since we are taking the ratio of the two local normalization distortions, the global constants C cancel out and so do not need computing.

For $k=5, 50, 150$, we start by finding values of $\pi$ and $\tau$ such that on average, for a randomly chosen context, ${\bf y_{<t}}$, $Z_k({\bf y_{<t}})\approx Z_{\pi}({\bf y_{<t}})\approx Z_{\tau}({\bf y_{<t}})$. We are seeking here to tune our parameters so that the average amount of renormalizing done by top-k, nucleus and temperature sampling is the same. 

Having tuned parameters, we then compare the local normalization distortion across the three decoding strategies. Starting with the single word context `The ', we generate $1000$ pairs of completions of $100$ tokens each for each of our decoding strategies and parameter choices. We compute the relative local normalization distortion given by the quantity (\ref{ratioLND}) for top-k sampling and equivalent quantities for nucleus and temperature sampling, see Appendix \ref{sec:computingLND}.

Our results are presented in Figure \ref{fig:LNDComparison} and Table \ref{table:quantiles}. We have two key findings.
\begin{figure*}[ht]\label{Fig2}
    \centering
\includegraphics[width=\linewidth]{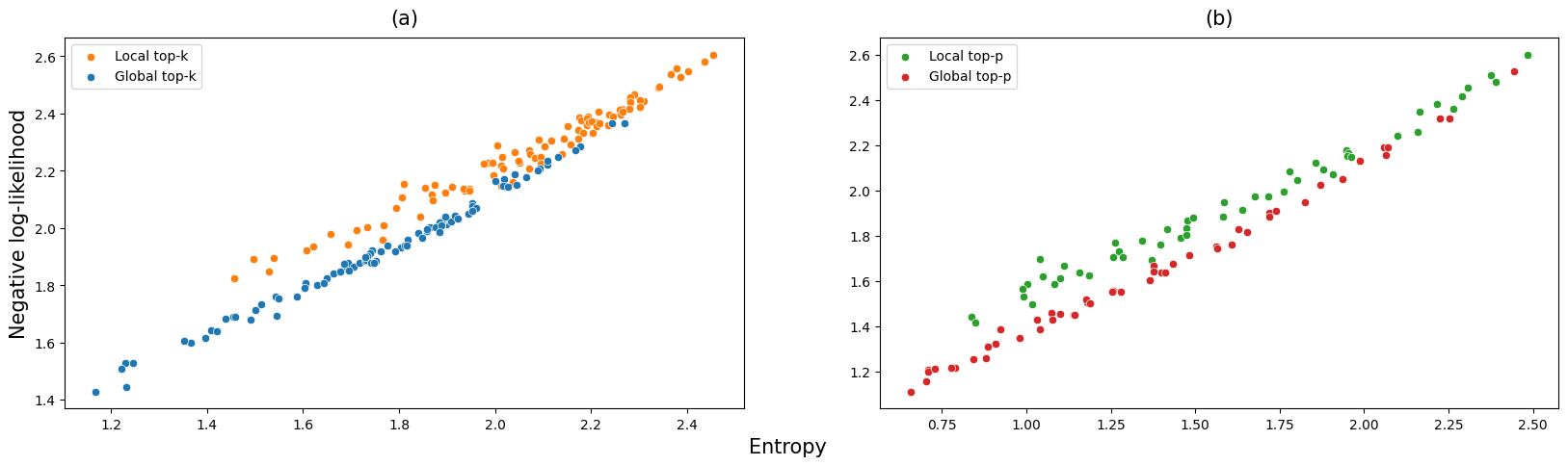}
    \caption{Evaluating quality against diversity at different parameter values when decoding with top-k, nucleus sampling, and globally normalized top-k. The ranges were $k=10,11, \dots,100$ and $p=0.4,0.41,\dots,0.9$. Higher values of entropy are better, lower values of negative-$\log$-likelihood are better. We find that at any fixed value of entropy, globally normalized sampling tends to produce texts with higher log-likelihood.}
    \label{fig:enter-label}
\end{figure*}
\begin{table}[h!]
\centering
\caption{Local normalization distortion ratios over a range of comparable decoding strategies. The table is partitioned into three groups of parameters tuned so that the average amount of renormalization is approximately equal. Reported quantiles are the absolute value of the natural log of the ratio (\ref{eq:LNDRatio}).}\label{table:quantiles}
\resizebox{\linewidth}{!}{%
\begin{tabular}{l*{5}{c}}
\toprule
\multicolumn{1}{c}{Decoding} & \multicolumn{5}{c}{Quantile} \\
 \cmidrule(lr){2-6}
\multicolumn{1}{c}{Strategy} & {10\%} & {25\%} & {50\%} & {75\%} & {90\%} \\
\midrule
$k=5$                        & 1.84          & 4.50          & 9.82          & 16.93         & 25.07          \\
$\tau=0.86$                  & 1.18          & 2.76          & 5.97          & 10.80         & 15.65          \\
$p=0.65$                     & \textbf{0.73} & \textbf{2.05} & \textbf{4.58} & \textbf{7.91} & \textbf{11.32} \\ \midrule
$k=50$                       & 0.60          & 1.43          & 3.26          & 5.60          & 7.92           \\
$\tau=0.95$                  & 0.40          & 1.05          & 2.29          & 3.93          & 5.83           \\
$p=0.88$                     & \textbf{0.18} & \textbf{0.44} & \textbf{0.95} & \textbf{1.76} & \textbf{2.70}  \\ \midrule
$k=150$                      & 0.36          & 0.88          & 1.89          & 3.35          & 4.83           \\
$\tau=0.98$                  & 0.18          & 0.47          & 0.95          & 1.68          & 2.43           \\
$p=0.95$                     & \textbf{0.06} & \textbf{0.16} & \textbf{0.34} & \textbf{0.62} & \textbf{0.96}  \\ \bottomrule
\end{tabular}}
\end{table}

\paragraph{Finding 1.} Local normalization distortion has a large effect. For example, when using temperature sampling with parameter $\tau=0.86$ to sample pairs of sequences $\bf w, \bf z$, each of length 100, we see that in half of cases, the ratio ${q_{\tau}({\bf w|c})}/{q_{\tau}({\bf z|c})}$ differs from the ratio ${p({\bf w|c})^{\frac{1}{\tau}}}/{p({\bf z|c})^{\frac{1}{\tau}}}$ by a factor of at least $\exp(5.97)\approx 392$. That is, the effect of local normalization distortion is to distort the relative probabilities of two completions by a factor of $392$.

It is worth stressing that temperature sampling is more often used with temperatures 0.7 or 0.8, in which case one would see even larger local normalization distortion.  

\paragraph{Finding 2.} When parameters $k,\tau$ and $\pi$ are tuned so that the typical renormalization factors $Z_k, Z_{\tau}$ and $Z_{\pi}$ are similar, nucleus sampling results in a much smaller local normalization distortion than temperature sampling, which in turn gives rise to a much smaller distortion than top-k sampling. 

\subsection{How does Local Normalization Distortion affect the Quality-Diversity Trade Off?}\label{sec:experiment7}

Given a single word context $w_1=\textnormal{`The '}$, we generate $30$ length $15$ samples by top-k sampling, nucleus sampling, temperature sampling and their globally normalized equivalents. We do this over a range of values of k, $\pi$ and $\tau$. For each sample $w_2\cdots w_{16}$, we assess the quality  by computing the negative log probability $-\log(p(w_2\cdots w_{16}|w_1))$. In addition, we evaluate the diversity of our sample generation process by approximating the entropy of our generation process using the Shannon-McMillan-Breimann theorem \cite{walters2000introduction}. That is, for each choice of decoding strategy and parameter value and each completion $w_2\cdots w_{16}$ generated by decoding strategy $q$, we compute $-\log(q(w_2\cdots w_{16}|w_1))$. We average these values over the different completions generated for each particular decoding strategy and parameter value to approximate $H(q)$. Figure \ref{fig:enter-label} shows these proxies for quality and diversity. Note that lower negative log probability and higher entropy are preferable.

\paragraph{Finding 3.} For both top-k and nucleus sampling, globally normalized sampling outperforms locally normalized sampling on the quality-diversity trade off. That is, at any fixed value of entropy, globally normalized sampling produces texts with higher log-likelihood. 
\newline\\
It is not tractable to compute globally normalized temperature sampling for sensible ranges of $\tau$, due to the extreme rejection rate of our rejection sampling algorithm. Fortunately, we do not require experimental results in this setting thanks to Corollary \ref{globaleqstate}, which states that globally normalized temperature sampling at temperature $\tau$ yields the unique distribution maximizing entropy plus ${1}/{\tau}$ log-likelihood among measures on on $\mathcal A^T$.

\section{Conclusions}
Our primary contribution is to express popular decoding strategies as equilibrium states. In doing so, one can see that the quantity which they maximize contains a term relating to local normalization of probability mass which seems unrelated to any reasonable goal of a decoding strategy. In particular, this term pulls the resulting probability distribution away from the quality-diversity maximizing curve. We have shown experimentally that the effect of local normalization distortion on the probability of selecting a string is typically very large (Section \ref{sec:experiment6}), and that it has a strongly negative effect on the quality-diversity tradeoff (Section \ref{sec:experiment7}) when these quantities are measured through entropy and log-likelihood. In a follow-up work \cite{kempton2025temptest} we build a detector of machine generated text based on local normalization distortion which outperforms state of the art alternatives. These factors lead us to the conclusion that local normalization distortion may have a negative effect on machine-generated text and that it should be carefully considered both when practitioners choose a decoding strategy and in the design of future methods for detecting machine-generated text.

\section{Ethical Considerations}

This work considers current decoding strategies for language models and ways in which these decoding strategies fall short. The most likely practical applications of it are in the detection of machine-generated text and in improving language models so as to make their outputs more human-like. 

Although there are no specific ethical concerns about this work, we do inherit wider ethical questions around building human-like language models and detecting machine-generated text. A discussion of these is far beyond the scope of this work; instead, we encourage the reader to seek out the wealth of publicly available material on the issues.

\section{Limitations}

Our experiments are run on the open-source Llama 2 language model \citep{touvron2023llama}. While this is not uncommon for research in computational linguistics, the setting in which decoding strategies such as temperature sampling are most widely deployed is closed-source models such as with ChatGPT \cite{openai2022chatgpt}. Our theoretical results hold for all language models and we do not believe the conclusions of our experimental section would change with language model. It is however the case that the magnitude of local normalization distortion would decrease if the entropies of the next token probability distributions were typically lower. Thus, language models with higher certainty about their next token predictions would give rise to smaller numbers in Table \ref{table:quantiles}, for example.

We have used log-likelihood as a proxy for the quality of machine-generated text and entropy as a proxy for its diversity. These metrics are clearly imperfect. In particular, human judgement of text quality may be a better metric for the quality of a text, although obtaining these human judgments is often prohibitive due to cost \cite{clark2021all}. 

\bibliography{NLPTemp}
\newpage

\appendix

\section{Computing Local Normalization Distortion for Nucleus and Temperature Sampling}\label{sec:computingLND}

In equations \ref{eq:LNDRatio} and \ref{ratioLND} we described how to compute the ratio of the local normalization distortions of two strings $y_m\cdots y_T$ and $z_m\cdots z_T$ in the case of top-k sampling. 

The case of nucleus sampling is almost identical, we need only replace $q_k$ with $q_{\pi}$ in equation \ref{ratioLND}.

For temperature sampling, we note that $q_{\tau}'(y_m\cdots y_T|{\bf y_{<m}})$ is not proportional to $p(y_m\cdots y_T|{\bf y_{<m}})$, but to $p^{\frac{1}{\tau}}(y_m\cdots y_T|{\bf y_{<m}})$. Thus, to compute the ratio of the local normalization distortions for two strings $y_m\cdots y_T$ and $z_m\cdots z_T$ in the case of temperature sampling, we replace equation \ref{ratioLND} with
\[
\dfrac{q_{\tau}(y_m\cdots y_T|{\bf y_{<m}})}{p^{\frac{1}{\tau}}(y_m\cdots y_T|{\bf y_{<m}})}. \dfrac{p^{\frac{1}{\tau}}(z_m\cdots z_T|{\bf y_{<m}})}{q_{\tau}(z_m\cdots z_T|{\bf y_{<m}})}.
\]

\section{Rejection sampling algorithms}\label{app:rejection}
We can sample from $q'_k, q'_{\pi}$ and $q_{\tau}'$ using rejection sampling. This remains incredibly computationally intensive, but it is significantly easier than computing the probability of each possible string $w_1\cdots w_T$ as in equation \eqref{eq:global_top_k}.

In the case of top-k and nucleus sampling, with set of allowed completions $\mathcal A^*_{{\bf y_{<m}}}$ given context ${\bf y_{<m}}$, one can sample according to the globally normalized variant of top-k or nucleus as follows:

\paragraph{Step 1.} Sample a completion $y_m\cdots y_T$ according to the model probability $p$.

\paragraph{Step 2.} Accept the completion $y_m\cdots y_T$ if it is in the allowed set $\mathcal A^*_{{\bf y_{<m}}}$, otherwise reject it and repeat step 1.

In the case of temperature sampling, given context ${\bf y_{<m}}$  one can sample according to a globally normalized variant of temperature sampling as follows:

\paragraph{Step 1.} Sample a completion $y_m\cdots y_T$ according to the model probability $p$.

\paragraph{Step 2.} Accept this completion with probability $p(y_m\cdots y_T|{\bf y_{<m}})^{\frac{1}{\tau}-1}$. If the completion is not accepted, return to step 1.

We see that with each attempt to sample according to globally normalized temperature sampling, sample $y_m\cdots y_T$ is generated with probability 
\begin{eqnarray*} & &p(y_m\cdots y_T|{\bf y_{<m}})\times p(y_m\cdots y_T|{\bf y_{<m}})^{\frac{1}{\tau}-1}\\
&=& p(y_m\cdots y_T|{\bf y_{<m}})^{\frac{1}{\tau}}.\end{eqnarray*}

For further information on rejection sampling and faster algorithms which approximate rejection sampling see \newcite{lipkin2025fastcontrolledgenerationlanguage}.

\section{Global Temperature Normalization in Ergodic Theory and Fractal Geometry}\label{sec:gibbsmeasures}
We mentioned in Section \ref{sec:globalnorm} that global normalization of measures is a standard method in statistical physics, ergodic theory and fractal geometry. A short explanation of this remark is that globally normalized temperature sampling corresponds to taking the Gibbs-equilibrium measure associated to potential $\log p$ at temperature $\frac{1}{\tau}$. Similarly, when using a truncation sampling algorithm with allowed set $\mathcal A$, globally normalized sampling corresponds to sampling from the Gibbs-equilibrium measure associated to potential $\log p$ on the sequence space defined by $\mathcal A$. For both of these comments, see \newcite{bowen2008equilibrium}.

For a more direct example, consider extremely long texts generated by pure sampling from a language model with finite context length $L$. The ergodic theory of Markov chains tells us that, with high probability, the average value of the log probability of a token from the text (`time average') will be close to the space average $$\int_{\mbox{contexts }v_{<t}}\int_{v\in\mathcal V} p(v|v_{<t}) dp(v|v_{<t})dp(v_{<t}).$$ 

One might ask what can be said about the set of texts for which the average log probability of tokens takes some different value $\alpha$. How are typical such texts distributed? How many are there? Such questions are answered through the multifractal analysis of ergodic averages, see, for example, \newcite{falconer2007fractal}. Solutions involve globally (rather than locally) normalized temperature sampling.
\section{Proofs}\label{sec:proofs}
\subsection{Proof of Proposition \ref{prop:zerotemp}}\label{proof:zerotemp}
We recall Proposition \ref{prop:zerotemp}, which stated that the limit as temperature tends to zero of locally normalized temperature sampling is greedy decoding, whereas the limit as temperature tends to zero of globally normalized temperature sampling is the distribution achieving globally maximal average log-likelihood.

The statement on local temperature sampling has been widely noted. It is merely the statement that for any probability vector $(p_1,\cdots,p_k)$, with a unique value $p_i$ larger than all other values, the vector 
\[
\left(\dfrac{p_1^{\frac{1}{\tau}}}{\sum_{j=1}^k p_j^{\frac{1}{\tau}}},\dfrac{p_2^{\frac{1}{\tau}}}{\sum_{j=1}^k p_j^{\frac{1}{\tau}}},\cdots , \dfrac{p_k^{\frac{1}{\tau}}}{\sum_{j=1}^k p_j^{\frac{1}{\tau}}}, \right)
\]
converges to the unit vector with a 1 in position $i$ as $\tau\to 0$.

The statement on global temperature sampling is more subtle and is a key result linking `zero temperature limits of Gibbs measures' and `ergodic optimization', see for example \newcite{bremont2002gibbs, jenkinson2019ergodic}.

\subsection{Proof of Corollary \ref{topkeqstate}.} We take Lemma \ref{lemma:bowen} and set $X=\mathcal A^*_{{\bf y_{<m}},k}$ to be the set of completions $y_m\cdots y_T$ belonging to the top-k set. Our distribution $q_k$ is a probability measure on this set. Then Lemma \ref{lemma:bowen} says that $q_k$ is the unique probability measure maximising the quantity
\[
H(\mu)+\sum_{{\bf w}\in \mathcal A^*_{{\bf y_{<m}},k}} \mu({\bf w|y_{<m}})\log q_k({\bf w|y_{<m}})
\]
among probability measures $\mu$ on the top-k set $\mathcal A^*_{{\bf y_{<m}},k}$. Note that 
\begin{align*}
&q_k({\bf w|y_{<m}})\\
=& \prod_{i=0}^{T-m}q_k(w_{m+i}|y_0\cdots y_mw_{m+1}\cdots w_{m+i-1})\\
=&\prod_{i=0}^{T-m}\dfrac{p(w_{m+i}|y_0\cdots y_mw_{m+1}\cdots w_{m+i-1})}{Z_k(y_0\cdots y_mw_{m+1}\cdots w_{m+i-1})}\\
=&\dfrac{p({\bf w|y_{<m}})}{\prod_{i=0}^{T-m}Z_k(y_0\cdots y_mw_{m+1}\cdots w_{m+i-1})}.
\end{align*}
Taking logs and splitting the formula for $\log q_k({\bf w|y_{<m}})$ into two distinct terms gives the result.

\subsection{Proofs of Corollaries \ref{toppeqstate}, \ref{tempeqstate} and \ref{globaleqstate}.} These corollaries follow from Lemma \ref{lemma:bowen} in an identical manner to the above proof of Corollary \ref{topkeqstate}.

\section{Supplementary Figures and Replication on other Language Models}\label{supp:app}
\begin{figure*}\label{LNDParameters}
    \centering
\includegraphics[width=\linewidth]{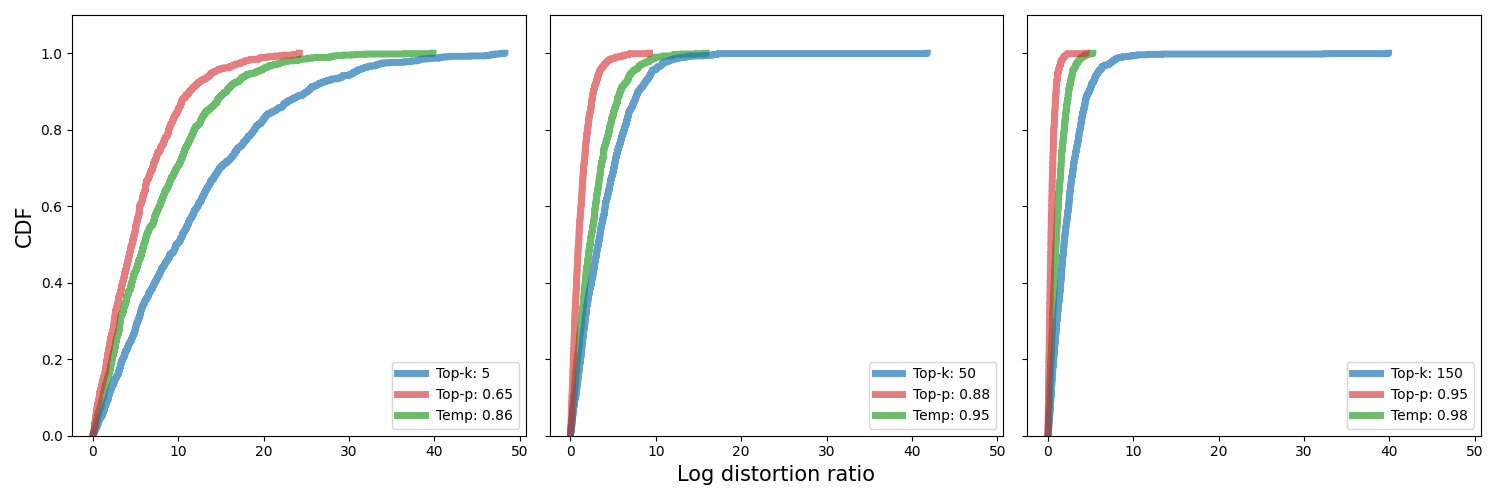}
    \caption{For each graph we generate pairs of texts according to top-k, nucleus or temperature sampling with Llama 2 7B. Parameters for the decoding strategies are tuned so that typical renormalizing quantitites $Z_k,$ $Z_{\pi}$ and $Z_{\tau}$ are of the same size. We then plot the log of the ratio of the local normalization distortion for the pair of generated texts, and then plot the cumulative distribution function. We see in each case that nucleus sampling produces the smallest local normalization distortion, followed by temperature sampling and top-k sampling.}
    \label{fig:LNDComparison}
\end{figure*}

Figure \ref{fig:LNDComparison} is a companion to Table \ref{table:quantiles} giving the results of Section \ref{sec:experiment6} across all quantiles. 

The experiments reported in the main body of the text were carried out using Llama 2 7B. We repeat these experiments here for other language models from the Llama and Pythia families, and find that results align. We tune values of p and $\tau$ to $k$ as described in Section \ref{sec:experiment6}. This process will have some noise and we note that the tuned values are slightly different for each model.

\begin{figure*}
    \centering
\includegraphics[width=\linewidth]{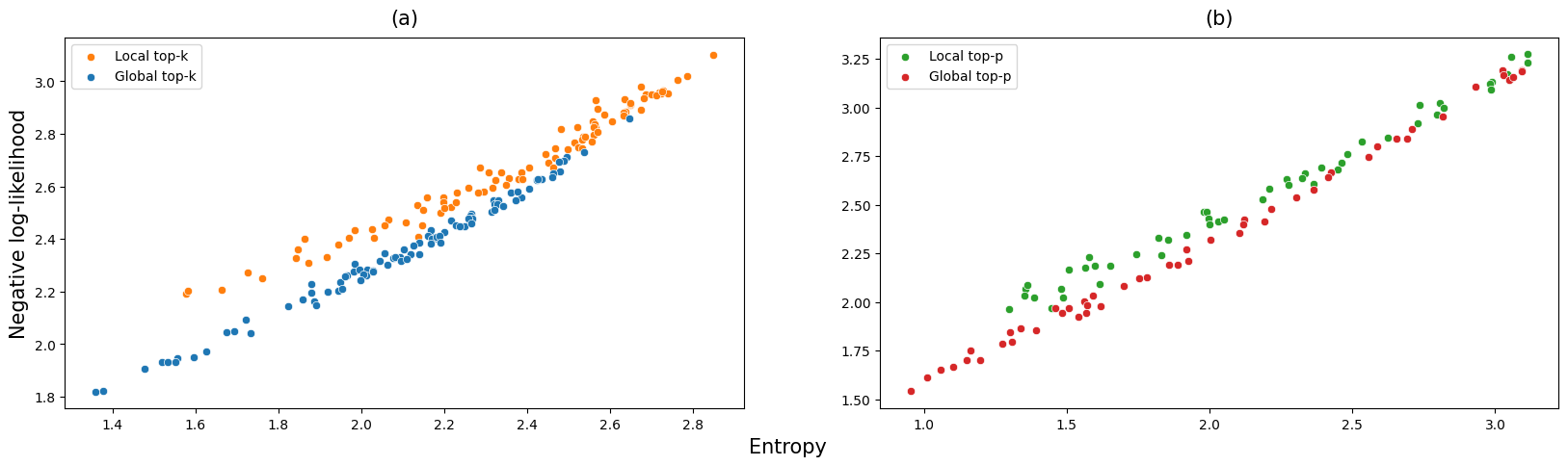}
    \caption{Figure \ref{fig:enter-label} repeated for Llama 3.2 1B.}
\end{figure*}

\begin{figure*}
    \centering
\includegraphics[width=\linewidth]{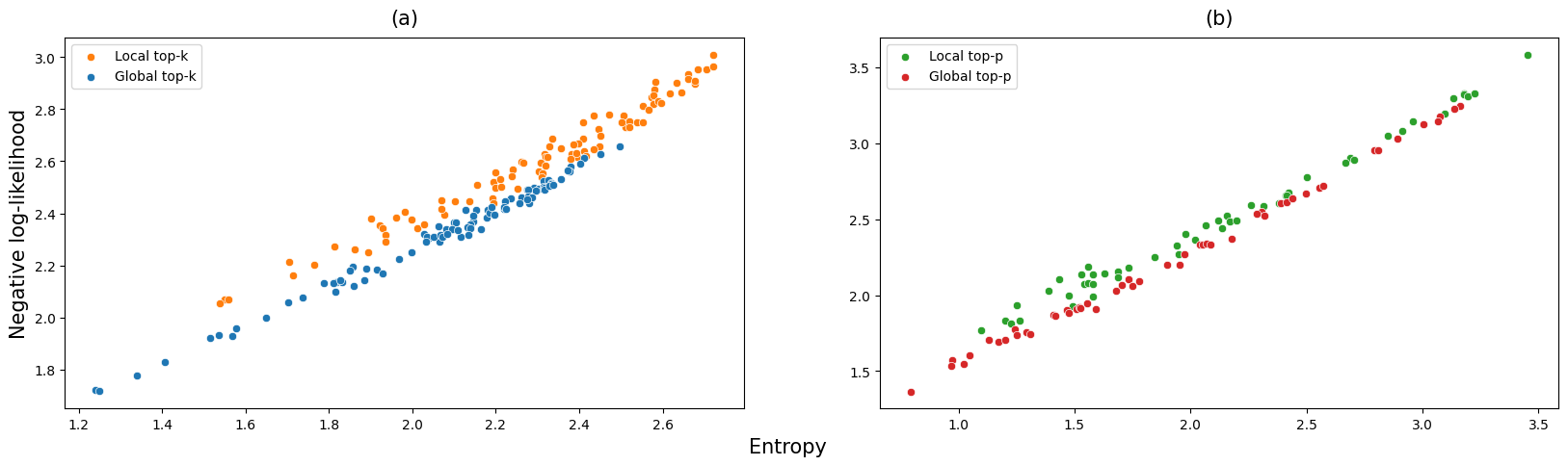}
    \caption{Figure \ref{fig:enter-label} repeated for Llama 3.2 3B.}
\end{figure*}

\begin{figure*}
    \centering
\includegraphics[width=\linewidth]{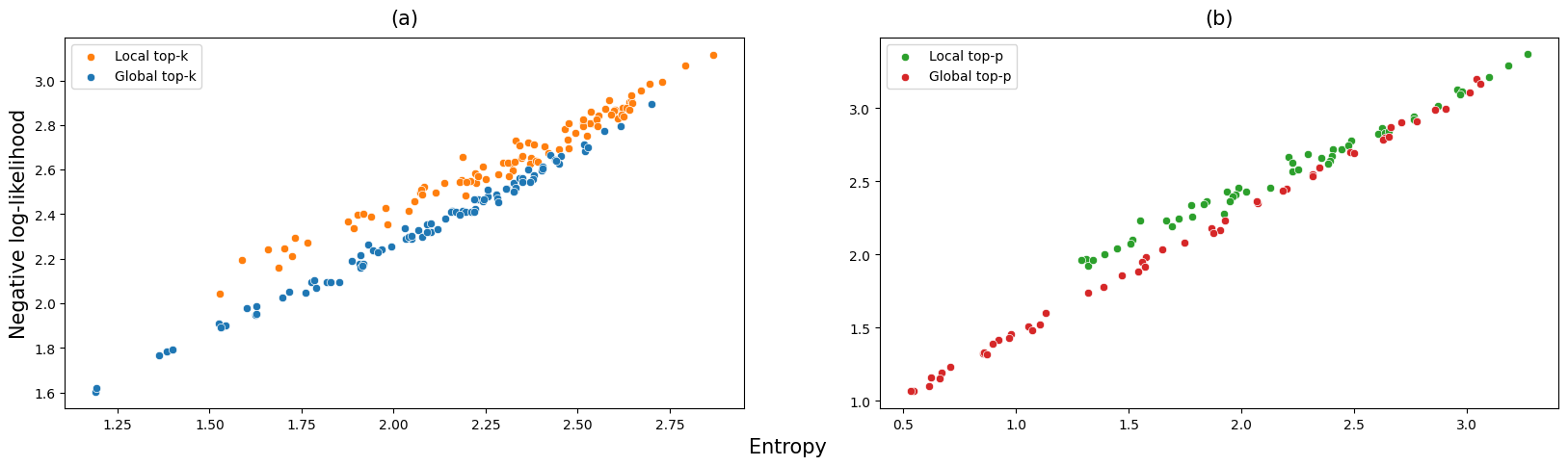}
    \caption{Figure \ref{fig:enter-label} repeated for Pythia 1B.}
\end{figure*}

\begin{figure*}
    \centering
\includegraphics[width=\linewidth]{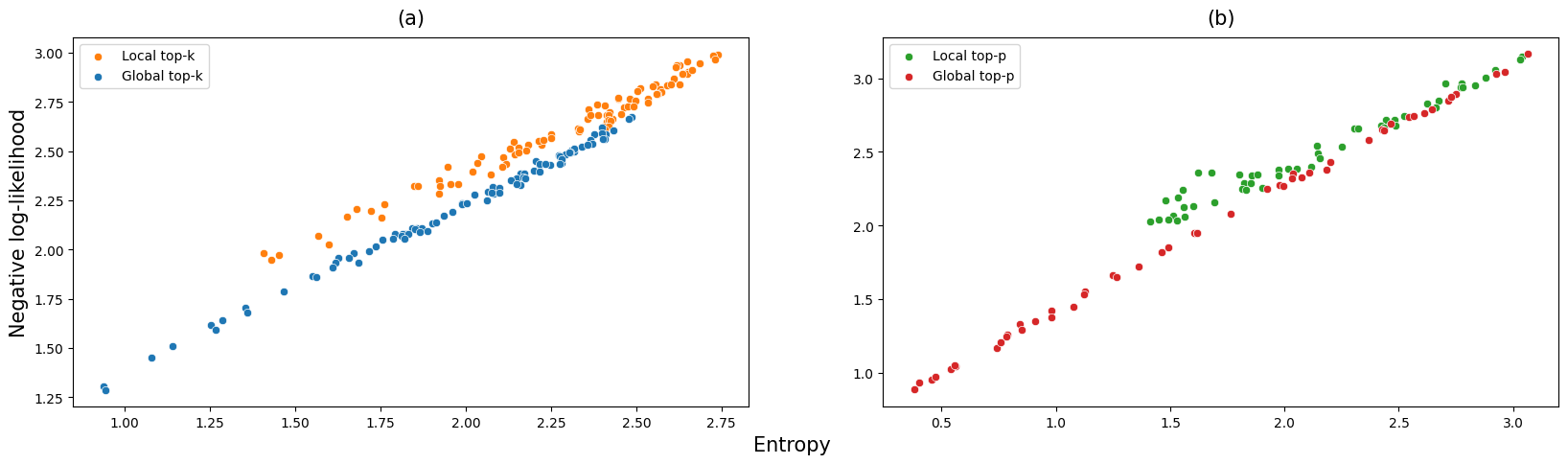}
    \caption{Figure \ref{fig:enter-label} repeated for Pythia 2.8B.}
\end{figure*}

\begin{figure*}
    \centering
\includegraphics[width=\linewidth]{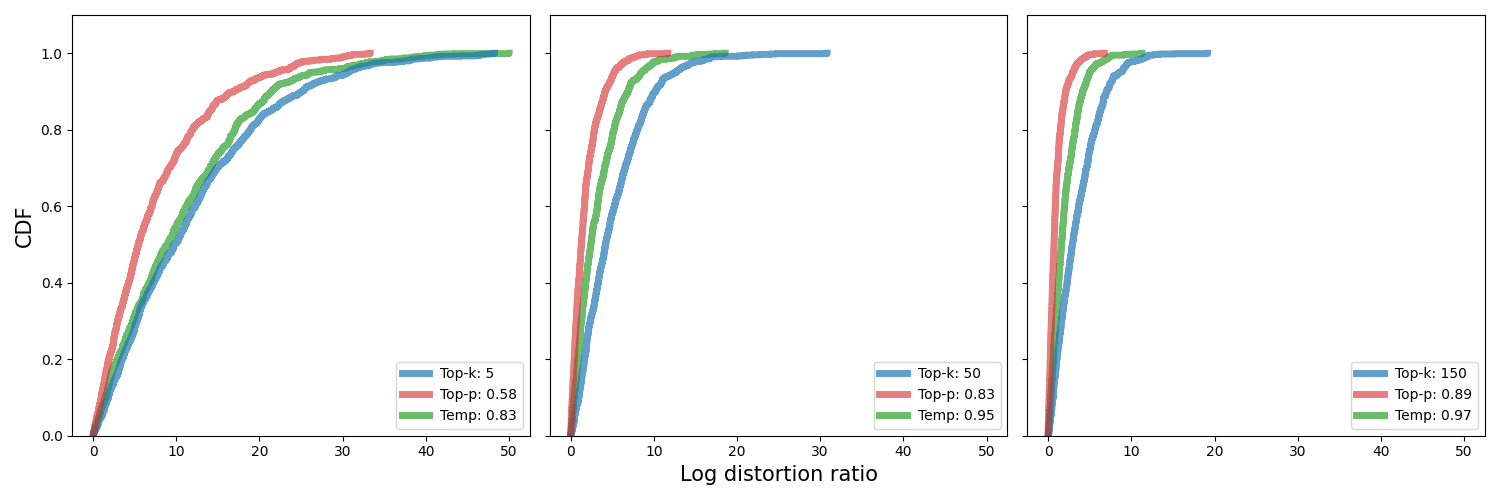}
    \caption{Figure \ref{fig:LNDComparison} repeated for Llama 3.2 1B.}
\end{figure*}
\begin{figure*}
    \centering
\includegraphics[width=\linewidth]{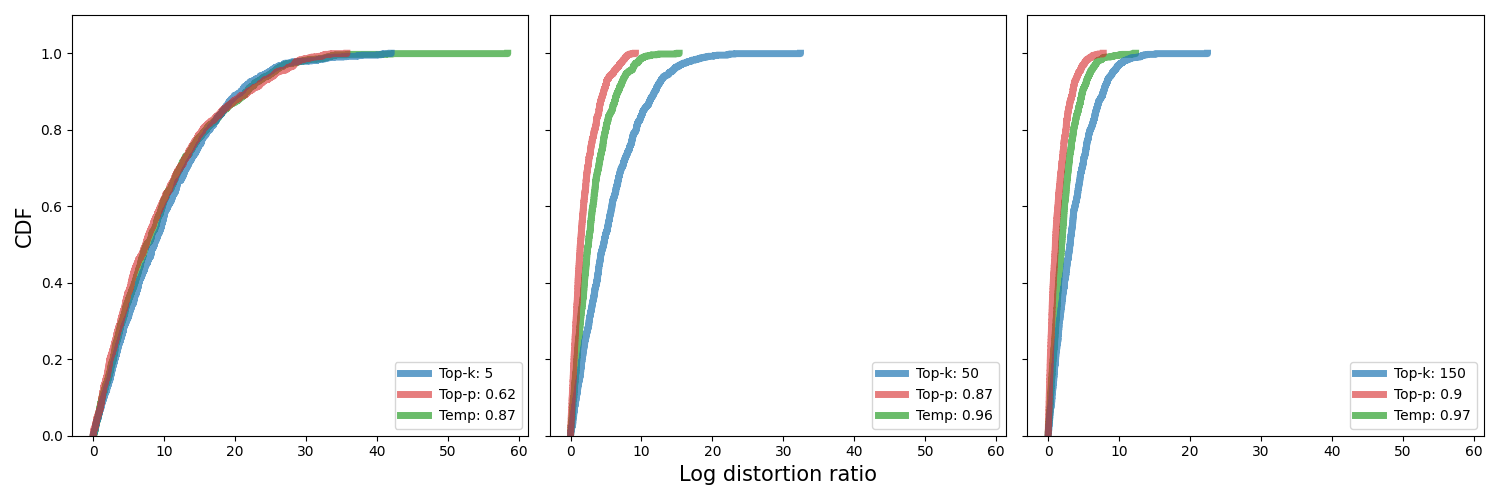}
    \caption{Figure \ref{fig:LNDComparison} repeated for Llama 3.2 3B.}
\end{figure*}

\begin{figure*}
    \centering
\includegraphics[width=\linewidth]{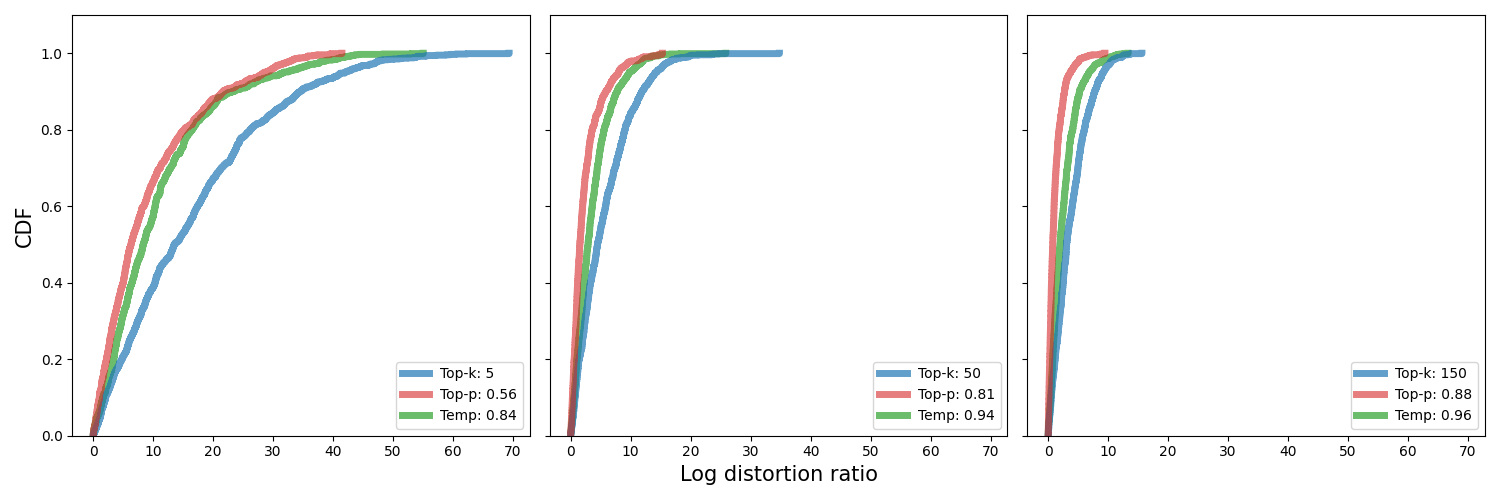}
    \caption{Figure \ref{fig:LNDComparison} repeated for Pythia 1B.}
\end{figure*}

\begin{figure*}
    \centering
\includegraphics[width=\linewidth]{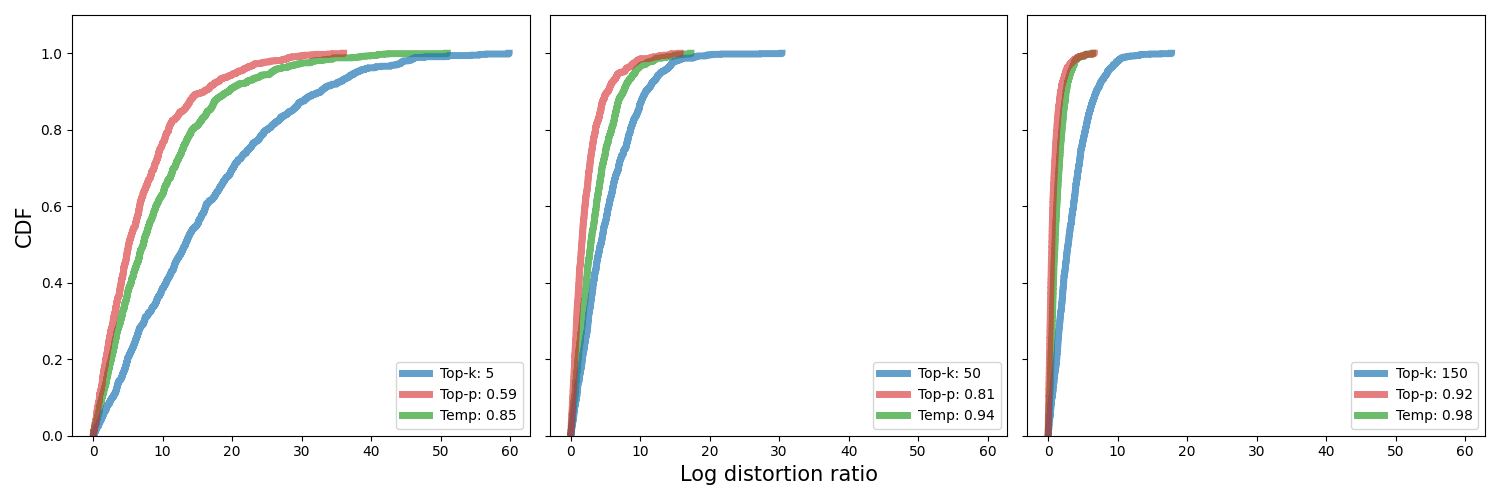}
    \caption{Figure \ref{fig:LNDComparison} repeated for Pythia 2.8B.}
\end{figure*}
\section{A Comment on Timescales for Normalization} In this article we have discussed two different timescales for renormalizing probabilities which do not sum to one, namely locally renormalizing conditional probabilities at each timestep and globally renormalizing probabilities of strings $w_1\cdots w_T$ at time T, where T is the length of the string we wish to generate. There is (at least in theory) a third natural option, which is to renormalize `at time infinity'. For example, in the case of temperature sampling at temperature $\tau$, this corresponds to taking the Gibbs measure associated to potential $p^{\frac{1}{\tau}}$ \cite{bowen2008equilibrium}. Precisely, there exists a unique measure $q_{\tau}''$ on the set $\mathcal A^{\infty}$ such that there exist constants $C,P$, independent of $T$, such that 
\begin{align*}
\frac{1}{C}&\leq \dfrac{q_{\tau}''(\{\underline z\in\mathcal A^{\infty}:z_1\cdots z_T=w_1\cdots w_T\})}{p^{\frac{1}{\tau}}(w_1\cdots w_T)\exp(T.P)}\\
&\leq C.
\end{align*}
The constant $P$, known as the topological pressure, could in theory be explicitly computed as the maximal eigenvalue of a very large matrix. 

Normalizing at time infinity has some theoretical appeal, in that it produces a Markov measure whose transition probabilities do not depend on the sequence length $T$.

\section{A Note on Quality and Diversity}
Quality and diversity are broad terms which could admit several interpretations. The notion that our metrics attempt to capture can be described as follows. Consider the scenario in which a language model is prompted `please tell me a joke'. In our interpretation, the model would be judged high on the quality metric even if it always responded with the same joke, provided the joke was a good one. The model would be judged high on the diversity metric provided there is a high chance that, when prompted twice, it would produce different outputs, even if these outputs are nonsense, or if an individual output repeats itself, or if the different outputs are semantically similar. 

\section{Comparisons with the Article: \newcite{gareev-etal-2024-local}}\label{sec:Gareev}
We would like to thank one of our anonymous referees for making us aware of the article \newcite{gareev-etal-2024-local}. Their headline conclusion, `in most configurations, global decoding performs worse than the local decoding versions of the same algorithms', seems in direct opposition to ours, while their results are actually entirely consistent with our own. We carefully point out how the apparent inconsistencies arise. 

\subsection{Parameter Pinning and Results on Diversity}
In comparing the effect of local and global decoding on (proxies for) quality and diversity, we follow the approach of \newcite{caccia2019} of plotting quality and diversity on the same plot (as in Figure \ref{fig:enter-label}). This allows us to compare, for example, locally normalized top-k vs globally normalized top-k across the whole range of parameters. \newcite{gareev-etal-2024-local} show that, for any fixed value of $k$, local decoding produces text which is {\bf more} diverse than global decoding. We show that, at any desired level of quality of output, local decoding produces text which is {\bf less} diverse than global decoding. There is no contradiction here and we agree with their result, as can be seen that the blue and red dots in Figure \ref{fig:enter-label} are generally to the left of the orange and green dots. The apparent discrepancy arises since, if one wishes to compare local and global top-k at some fixed threshold for quality, one needs to use different values of k for the local and global decoding.

\subsection{Results on Quality}
As with diversity, \newcite{gareev-etal-2024-local} plot quality against parameter for local and global decoding, rather than plotting quality and diversity. However, in the case of their results on quality, we do not think this is the cause of the apparent discrepancy between our results and theirs. Instead, we think there are two fundamental factors.

MAUVE, their measure of quality, works by measuring both type 1 errors, where a language model produces un-human like text, and type-2 errors, where a language model fails to capture the full diversity of human language. This does not align with our desired notion of quality, which relates only to type-1 errors, see Section \ref{sec:evaluating}. Instead, in our language, MAUVE measures a convex combination of quality and diversity, and as such there is no clear discrepancy between their results and ours.  

We suspect however that a much larger factor is at play. In our experiments we produce texts of constant length. \newcite{gareev-etal-2024-local} have a rather clever way of approximately sampling from the globally normalized distribution, which allows them to produce much longer texts. A consequence of not requiring fixed length generation is that their globally sampled texts are much shorter than their locally sampled ones, for some parameters by a factor of nearly 4. This length discrepancy is the first of their three suggested explanations of their results. The fact that global and local sampling produce outputs of such starkly differing lengths is very interesting, but isn't really what we were hoping to measure when we talk about quality. Indeed, the fact that our generated texts are of constant length has allowed us to avoid the thorny question of whether we should be measuring quality, or quality per token, and we hope that further research might look at how our theoretical results would be affected if one were to normalize by generation length.

\section{Licenses}
We have used Llama 2, Llama 3, and Pythia models under their respective community license agreements.\footnote{See \url{https://huggingface.co/meta-llama} and \url{https://huggingface.co/EleutherAI}.}. Use for research is consistent with the terms of these licenses.

\end{document}